%% file: main.tex
\documentclass[sigconf]{acmart}

\usepackage{mathrsfs}	

\usepackage[subrefformat=parens,labelformat=parens]{subfig} 
\usepackage{confcopyright} 

\AtBeginDocument{%
  \providecommand\BibTeX{{%
    \normalfont B\kern-0.5em{\scshape i\kern-0.25em b}\kern-0.8em\TeX}}}

\setcopyright{acmcopyright}

\copyrightyear{2019}
\acmYear{2019}
\acmConference[SIGSPATIAL '19]{27th ACM SIGSPATIAL International Conference on Advances in Geographic Information Systems}{November 5--8, 2019}{Chicago, IL, USA}
\acmBooktitle{27th ACM SIGSPATIAL International Conference on Advances in Geographic Information Systems (SIGSPATIAL '19), November 5--8, 2019, Chicago, IL, USA}
\acmPrice{15.00}
\acmDOI{10.1145/3347146.3359358}
\acmISBN{978-1-4503-6909-1/19/11}

\begin{document}

\title{Bayesian Surprise in Indoor Environments}


\author{Sebastian Feld}
\affiliation{\institution{Mobile and Distributed Systems Group\\LMU Munich}}
\email{sebastian.feld@ifi.lmu.de}

\author{Andreas Sedlmeier}
\affiliation{\institution{Mobile and Distributed Systems Group\\LMU Munich}}
\email{andreas.sedlmeier@ifi.lmu.de}

\author{Markus Friedrich}
\affiliation{\institution{Mobile and Distributed Systems Group\\LMU Munich}}
\email{markus.friedrich@ifi.lmu.de}

\author{Jan Franz}
\affiliation{\institution{Mobile and Distributed Systems Group\\LMU Munich}}
\email{janfranz5993@gmail.com}

\author{Lenz Belzner}
\affiliation{\institution{MaibornWolff\\Munich}}
\email{lenz.belzner@maibornwolff.de}

\renewcommand{\shortauthors}{Feld et al.}

\begin{abstract}
This paper proposes a novel method to identify unexpected structures in 2D floor plans using the concept of Bayesian Surprise.
Taking into account that a person's expectation is an important aspect of the perception of space, we exploit the theory of Bayesian Surprise to robustly model expectation and thus surprise in the context of building structures. 
We use Isovist Analysis, which is a popular space syntax technique, to turn qualitative object attributes into quantitative environmental information.
Since isovists are location-specific patterns of visibility, a sequence of isovists describes the spatial perception during a movement along multiple points in space.
We then use Bayesian Surprise in a feature space consisting of these isovist readings.
To demonstrate the suitability of our approach, we take ``snapshots'' of an agent's local environment to provide a short list of images that characterize a traversed trajectory through a 2D indoor environment. 
Those fingerprints represent surprising regions of a tour, characterize the traversed map and enable indoor LBS to focus more on important regions.
Given this idea, we propose to use surprise as a new dimension of context in indoor location-based services (LBS). 
Agents of LBS, such as mobile robots or non-player characters in computer games, may use the context ``surprise'' to focus more on important regions of a map for a better use or understanding of the floor plan.

\end{abstract}

\ACMCopyright{2019}{
"Proceedings of the 27th ACM SIGSPATIAL International Conference on Advances in Geographic Information Systems ", 
Association for Computing Machinery,
New York,
NY,
USA, 
p. 129-138
}{10.1145/3347146.3359358}



\maketitle

\input{sections/introduction}
\input{sections/relatedwork}
\input{sections/background}
\input{sections/concept}

\input{sections/evaluation}

\input{sections/conclusion}

\bibliographystyle{ACM-Reference-Format}
\bibliography{main}

\end{document}

%% file: sections/introduction.tex
\section{Introduction}
\label{sec:introduction}

Location-based services (LBS) take advantage of a known location to process and provide information associated with it \cite{Kuepper2005}. 
A simple exemplary application is a search engine that displays results related to a given postal code. 
This basic principle can be refined with an automated location estimation, so that the user is always provided with information corresponding to the current location, such as all bus stops in the immediate vicinity.

A similar task is the navigation in buildings, hereafter called \emph{indoor navigation} \cite{Werner2014}. 
Again, there is an attempt to inform the user on the basis of the physical location in order to help finding the way around the building in a reliable manner. 
A very simple example of such an LBS is a floor plan containing a ``current location'' marker.
When designing such a system, an important factor comes into play: the perception of the spatial structure and the impression the person gets while navigating through it. 
For each view of a person at a time, it is possible to describe the corresponding field of view (FoV). 
The FoV can potentially be infinite but is usually limited by obstacles, such as walls. 
So-called \emph{isovists} \cite{Benedikt1979} define the three-dimensional, visible space of a given FoV. 
Specific isovist-based measures include various properties, such as the area or the circumference of the isovist \cite{Freksa2005}.

In this work, we consider an agent's expectation as a decisive factor for the perception of space, or the exact contradiction of that expectation.
An example is the steady traversion of a monotonous narrow passage and the sudden ending in a hall. 
Since the FoV can be measured with help of isovists, we use them as a base for the description of expectation.
In order to derive a mathematical model for expectation, we adapt the concept of \emph{Bayesian Surprise} as introduced in \cite{Baldi_Itti10nn}. 
The authors propose that the subjective expectation of a person can be defined as a conditional probability distribution of events and that this distribution is constantly updated and thus ``corrected'' by new measurements -- in our domain of application new FoWs described by isovist measures over time.


The aim of this work is to create a meaningful and well-defined link between the concept of Bayesian Surprise and indoor navigation.
Specifically, this is done using isovist measures along spatial trajectories. 
From this combination, both, recurrent structures and those that are contrary to the expectation -- thus surprising -- are reliably recognized along a route.

The motivating idea behind is to provide this novel kind of context to diverse domains of mobile agents. In a reinforcement learning scenario there could be a guided exploration via Bayesian Surprise, i.e. an exploring agent would try to maximize the surprise in the aspiration of learning new facts. Another domain would be safety in Industry 4.0, where mobile robots in a changing environment may react on dynamic situations using a surprise map, i.e. a surprised agent may state that the following actions are based on uncertainty. Lastly, an agent may use regions with high surprise to perform subgoal detection, thus splitting the map/trajectory at that particular parts.


The contribution of this paper can be divided into two main parts:
First, we present the novel combination of Bayesian Surprise with a description of spatial perception using isovist analysis.
In addition to the investigation and formulation of a suitable description of expectation, this also includes the combination or selection of individual factors that were used.
For this purpose, different modeling concepts in relation to the Bayesian Surprise were designed and evaluated.
Second, the proposed approach was implemented in the framework ``Unity'' by Unity Technologies in order to test, evaluate, and customize the application. 
In a virtual environment, building plans are visualized, with the actual computation of the surprise value being kept in two dimensions in order to reduce complexity. 
Since in this context, the human surprise can only occur through a change in the perceived environment, a huge number of routes are defined in the aforementioned floor plans. 
Routes include both, round trips and various ways to get from one point to another. 
Isovists are computed along these trajectories and their measured quantities are determined.
The resulting isovist properties are evaluated with the help of the adapted concept of Bayesian Surprise, whereby the recognition of recurrent structures as well as the recognition of unexpected structures served as focal points.


The paper is structured as follows: Section \ref{sec:related_work} provides an overview of similar and related research, whereas Section \ref{sec:background} introduces the most important concepts and definitions used throughout the paper.
This is followed by Section \ref{sec:concept} which details the proposed approach whose evaluation is part of Section \ref{sec:evaluation}. 
The paper concludes with some thoughts of future work in Section \ref{sec:discussion}.  

%% file: sections/relatedwork.tex
\section{Related Work}
\label{sec:related_work}
Regarding the topic of this paper, several elaborations exist that cover similar scientific areas and some of their findings were used in this work. 
The core is \cite{Baldi_Itti10nn}, where the ``Bayesian Theory of Surprise''\footnote{\url{http://ilab.usc.edu/surprise/}} is formulated, which is a mathematical definition of surprise.
In the following, we will describe related work in the field of spatial impression and research in applications of Bayesian Surprise.


The application of spatial impression is often based on the calculation and use of isovists. 
For example, \cite{Benedikt1979} states that the collection of all visible points from a given vantage point (= isovist) is relevant for behavioral and perceptual studies -- two categories, to which the term \emph{surprise} can be assigned to.
The paper \cite{feld2016approximated} describes an approach to approximately compute discrete isovists by performing ray casts.
The authors use isovists to semantically evaluate floor plans. 
For the paper at hand, we adopted the proposed ray casting technique.
\cite{feld2016identifying} also deals with the evaluation of spatial impressions using isovists. 
Here, rooms and areas of the floor plans are grouped according to their isovist properties. 
Although this approach allows for the definition and grouping of ways to a goal that are strongly different in terms of their perception, it does not consider the agent's expectation.
Closely related to our work is \cite{sedlmeier2018discovering}, in which a Unity framework for the computation of 2D isovists in a three-dimensional environment was implemented. 
The authors showed that repetitive structures are also reflected in the isovist measurements -- a finding that led to the basic idea of our paper.
Furthermore, we used the aforementioned framework for the evaluation of our approach.


The theorem of \emph{Bayesian Surprise} has already been used in attention analysis and image recognition:
in \cite{Boehnke_etal11ejn} an analysis of visual neurons using Bayesian Surprise was conducted that corroborated the hypothesis that repeated stimuli lead to a decreased response of those neurons.
Furthermore, research exists on what attracts human attention in natural environments \cite{Itti_Baldi06cvnms, Mundhenk_etal09vr}.
There, surprise as a factor exceeded all other tested metrics.
Another paper deals with the detection of events in dynamic natural environments \cite{Voorhies_etal10vss}. 
Like our approach, aforementioned works are based on a continuous flow of visual data, but they differ in details: 
we consider building structures instead of natural environments. 
In addition, especially \cite{Voorhies_etal10vss} focuses more on the temporal component of the surprise.
Likewise, \cite{Einhaeuser_etal07jov} is to be mentioned, in which the conclusion is drawn that the human recognition of well-known scenes is significantly affected by surprise.
In this direction also points \cite{Itti_Baldi05cvpr}, which is based on the ``Bayesian Theory of Surprise'' as used also in our approach. 
Their basic idea is to find positions or sequences in videos that are considered unexpected by the human viewer. 
This has a close connection to our work, since we also use a continuous flow of visual data as well as the concept of Bayesian Surprise.
Finally, the authors of \cite{Itti_Baldi09vr} state that the general views of people are focused on spots that are classified by the Bayesian Surprise as astonishing. 
Conversely, it can be concluded that such surprising elements attract human attention which led to the basic idea of this work:
according to this model, surprising regions are ideal locations for information to be disseminated, for example to install position plans or other, more sophisticated location-based services built for indoor navigation.

%% file: sections/background.tex
\section{Background}
\label{sec:background}

For the general understanding of this work, it is necessary to be able to assign the terms Isovist Analysis and Bayesian Surprise together with the Kullback-Leibler Divergence.

\subsection{Isovist Analysis}
\label{subsec:isovist_analysis}

A single isovist itself is the formative volume of visible space from a given point of view \cite{Benedikt1979}. This ``shape affiliation'' is expressed by the fact that the isovist is bounded by the obstructive objects. This shape may or may not depend on the current location of the measurement. Thus, in a convex, closed and empty space, it can be assumed that the isovist remains unchanged from a displacement of the position. An easy way to visualize this is to describe the isovist as the light cone of a luminous sphere.

By nature, isovists are initially three-dimensional, but it is also possible to consider a two-dimensional cross-section. This may be imaginable in all angles, but in terms of this paper's domain -- indoor navigation -- only a horizontal cut is reasonable.

\begin{figure}[hpbt]
  \centering
  \includegraphics[width=.85\linewidth]{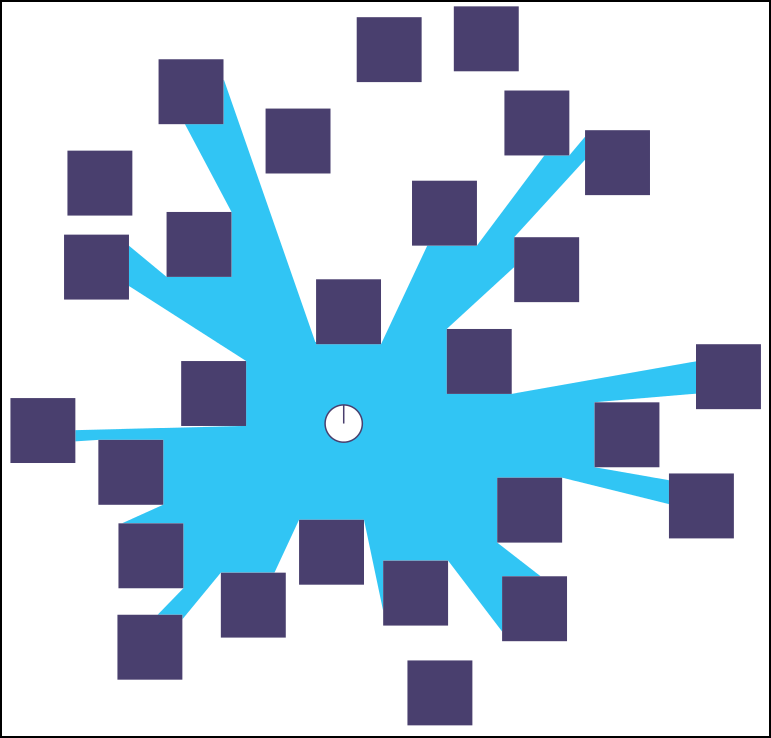}\\
  \caption{An exemplary 2D Isovist.}\label{fig:isovist_example}
\end{figure}

An example can be found in Figure \ref{fig:isovist_example} where the visible space is measured from the white circle while being bounded by the surrounding obstacles in the form of squares. This example also clarifies that an isovist is limited but not defined by the outer shape. An agent standing in place of the circle is by no means completely enclosed, but the resulting isovist is not infinitely large.

Isovist measures refer to evaluations of the calculated shape. The features used in this work are explained in more detail below.

Given a Euclidean three-dimensional space $E^3$ and a simple, connected area $D$ with its boundary line $\partial D$. One can think of $D$ as a building plan with $\partial D$ as the boundaries of the map (the blank white area in Fig. \ref{fig:isovist_example} together with the black borderline). Now, connected, substantial and visible points are defined as the ``real surface'' $S$. This generally refers to obstacles and walls, in Fig. \ref{fig:isovist_example} this would be the dark squares.
The isovist $V_x$ at viewpoint $x$ is now defined as $V_x = \left\{v \in D: v \text{ is visible from } x \right\}$, that is, all points in $D$ that are visible from $x$. The isovist's boundary $\partial V_x$ can then be divided into two parts: the real surface $S_x$ (boundary line running on obstacles or map borders) and the hidden radial boundary line $R_x$ (imaginary line of sight, which starts at the edges of obstacles and ends at obstacles or the map's boundary).
Another definition of isovists, according to Benedikt \cite{Benedikt1979}, is that an isovist can also be considered as a set of lines connecting the viewpoint $x$ with the points $v'$ on the visible boundary line $S_x$ of the isovist. It follows $V_x = \left\{ \left[ x, v' \right] : v' \in S_x \right\}$. These lines are below referred to as rays and have the length $l_{x,\theta} = d\left( x, v' \right) = \left\| v' - x \right\|$.

The definition of the six isovist measures are now as follows: The \textbf{area} is the cross section of the calculated isovist and defined by $A_x = A(V_x)$. This corresponds to the blue marked area in Figure \ref{fig:isovist_example}.
The \textbf{perimeter} $\partial V_x$ indicates the total edge length of the surface and is therefore only conditionally dependent on it, since complex tilted shapes can arise. The total edge length can be divided into two categories: the real-surface perimeter $P_x$ and the occlusion $Q_x$.
The \textbf{real-surface perimeter} $P_x = |S_x|$ indicates how much of the edge length actually runs along surfaces. In the given example these are the contact edges of the blue isovist with the darker obstacles.
In contrast, the \textbf{occlusion} $Q_x = |R_x|$ specifies the total length of all occluding edges. This is not to be equated with the wall surface obscured by it, rather it is the air lines that build up the boundaries between the isovist and the white surroundings.
Altough the \textbf{mean} is initially not defined as an actual isovist property, it is needed for further calculation. In our case it indicates the average distance of an observer to the walls. The length of the rays emitted to calculate the isovists are measured until their first collision.
Similar to the calculation of the mean value, the \textbf{variance} of the rays' length is used as an isovist measure, thus $M_{2,x} = M_2(l_x,\theta)$ is the second central moment with respect to the rays' length and describes the deviation from the arithmetic mean of the lengths.
The \textbf{skewness} refers to the direction and strength of the asymmetry of the distribution of the rays mentioned and is defined by $M_{3,x} = M_3(l_x,\theta)$.
The last value used is \textbf{circularity} which indicates in this context how round a room is. This is calculated by putting in proportion the mean as the expected area and the actual area: $N_x = {|\partial V_x|}^2 / 4 \pi A_x$.

\subsection{Bayesian Surprise}
\label{subsec:bayesian_surprise}


The basis of \emph{Bayesian Surprise} is Bayes' theorem \cite{Bayes1763}. This mathematical concept describes the calculation of conditional probabilities and the general formular reads:
\begin{align*}
&P(A|B) = \frac{P(B|A) * P(A)}{P(B)}
\end{align*}
Here, the individual components say the following: $P(A)$ is the probability that event $A$ occured, while $P(B)$ is the probability that event $B$ has been observed, both independently of each other. $P(A|B)$ is a conditional probability, i.e. the probability to observe event $A$ based on the condition that event $B$ occured. Correspondingly, $P(B|A)$ is the conditional probability that event $B$ occurs based on the condition that event $A$ has been observed.
In a sense, this is the inverse of conclusions. If $P(B|A)$ is known, $P(A|B)$ can also be inferred using this theorem -- as long as $P(A)$ and $P(B)$ are known.


The concept of \emph{Bayesian Surprise} was established as a means of calculating surprise values \cite{Baldi_Itti10nn}. This defines that a person's expectation of a particular event can be subdivided into different models $M\in \mathscr{M}$, each representing a possible outcome. Mathematically, the complete expectation can therefore be represented as follows:
\begin{align*}
&\mathscr{M} = \operatorname{Expectations\, or\, Model Space}\\
&\{P(M)\}M\in \mathscr{M}
\end{align*}
The recalculation of these imaginations when taking into account the newly measured data $D$ is defined by Bayes' theorem:
\begin{align*}
&\forall M\in \mathscr{M}, P(M|D)= \frac{P(D|M)}{P(D)} * P(M)
\end{align*}
The surprise to be determined is defined in the concept of Bayesian Surprise as the difference between the prior distribution of the expectation values and the posterior distribution. According to the concept of Bayesian surprise, the Kullback-Leibler divergence (KL-divergence) \cite{KL-Divergence} is most appropriate for the determination of the difference \cite{Itti_Baldi09vr}. The resulting formula is then:
\begin{align*}
& S(D,\mathscr{M}) = KL(P(M|D),P(M)) = \int_{\mathscr{M}}P(M|D)*log(\frac{P(M|D)}{P(M)}* dM)
\end{align*}


The just-mentioned KL-divergence defines a measure of the difference between two probability distributions. It has its origins in information theory, whose main goal is to measure the amount of information in a dataset. In this context it describes the resulting loss of information if a distribution would be mapped to another approximate distribution.
In the field of information theory, the entropy of a discrete random variable $X$ with possible values $x_{1},\ldots ,x_{n}$ and probability mass function $P(X)$ is defined as:
\begin{align*}
& H = -\sum_{i=1}^{N} P(x_i) * log(P(x_i))
\end{align*}
The KL-divergence is only a minor modification of this formula. To calculate the difference between the distributions $P$ and $Q$, the formula is adjusted as follows:
\begin{align*}
& D_{KL}(P||Q) = -\sum_{i=1}^{N} P(x_i) * (log(P(x_i)-log(Q(x_i))\\
&\qquad\qquad\ = -\sum_{i=1}^{N} P(x_i) * log(\frac{P(x_i)}{Q(x_i)})
\end{align*}

%% file: sections/concept.tex
\section{Concept}
\label{sec:concept}

This section is divided into two parts:
(1) we describe the framework used to generate isovist measures along spatial trajectories.
These trajectories are computed in a floorplan-based simulation environment and provide
the input for the following step
(2), the computation of Bayesian Surprise measures along the trajectories, i.e. posterior inference and KL-divergence calculation.

\subsection{Isovist Input Generation}
\label{subsec:isovist_input_generation}
\begin{figure}[htb]
    \centering
    \includegraphics[width=.49\textwidth]{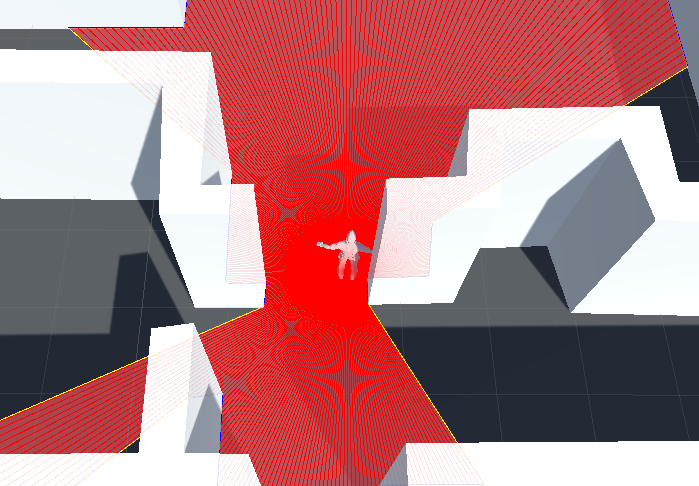}
    \caption{3D rendering in Unity showing the non-player character casting 360 rays (red lines) from it's current position.}
    \label{fig:concept_tum_rays}
\end{figure}

We make use of the framework for isovist generation on indoor floorplans as published in \cite{sedlmeier2018discovering}.
Apart from the already available maps published with the framework, which are based on floorplans of real buildings,
several new maps based on synthetic floorplans were developed for the work at hand.
These synthetic floorplans were designed for the goal of showcasing several basic concepts of surprise based on the perception of indoor space.
Note that the maps based on the real buildings do not contain doors, while the synthetic maps contain simulated doors, which are always closed.
While it is possible to move through these doors, it is impossible to look through them, i.e. they are non-translucent.
A more detailed analysis of the layout of the floorplans and the respective concepts is presented in Section \ref{sec:evaluation}.
Note that all maps used are generated by extruding 2D floorplans, so that although the maps are realized in three dimensions, only
2D isovists are calculated.
These maps serve as the basic asset for Unity, a 3D game engine and development environment \cite{unity}.
Navigation meshes are generated for each map to enable automatic navigation and pathfinding in Unity.
This allows a non-player character (NPC) to autonomously navigate the maps and find paths towards designated goalpoints on the maps.
These goalpoints can either be generated randomly or placed statically at fixed locations to showcase certain aspects on the synthetic maps.
When the simulation is run, the framework performs isovist calculation and logs the results to disk.
For every step of the NPC, a configurable amount of rays is cast from the current position.
Intersections of the rays with the map's mesh colliders then provide the necessary hitpoints for isovist measure calculation.
An example of how this looks like when visualized in Unity can be seen in Figure \ref{fig:concept_tum_rays}.

The isovist measures calculated are: real-surface perimeter, occlusion, area, variance, skewness and circularity, as described in Section \ref{subsec:isovist_analysis}.
For a detailed treatment of how these measures are calculated in Unity, see \cite{sedlmeier2018journal}.

\subsection{Calculation of Bayesian Surprise}
\label{subsec:calculation_of_bayesian_surprise}
As described in Section \ref{subsec:bayesian_surprise},
the theoretical formulation of Bayesian Surprise is agnostic to which specific types of distributions are used.
Consequently, the choice of which types of distributions to use for modelling the system has to be made use-case dependent.
When aiming to perform Bayesian inference, it is important to consider that the choice of distribution has an effect on the computational complexity of the calculation. 
In order to be able to compute the posterior analytically, conjugate priors can be used.
The interesting aspect when using conjugate priors is that the posterior belongs to the same functional family as the prior.
For the work at hand, we chose to model our data using separate Multinomial distributions per feature.
In other words, we assume that for each feature, the data $D$ follows a Multinomial distribution and consequently choose the functional form of $P(M)$ in such a way that $P(M|D)$ has the same form.
It can be shown that the correct form for $P(M)$ in this case is the Dirichlet distribution with probability density:
\begin{align*}
    P(M(x)) &= \frac{1}{B(\alpha)}\prod^K_{i=1}x^{\alpha_i-1}_i,
\end{align*}
with
\begin{align*}
    B(\alpha) &= \frac{\prod^K_{i=1}\Gamma(\alpha_i)}{\Gamma(\sum^K_{i=1}\alpha_i)}
\end{align*}
and concentration parameters $\alpha_1,...,\alpha_K > 0$, $K$ being the number of categories and $\Gamma(.)$ the Gamma function.

As the Multinomial distribution is a discrete probability distribution, while the isovist measures are continuous features, it is necessary to discretize the data.
For this, we assign each feature's data to linearly spaced, fixed amount of $k$ bins $b$, with bin $b_1$ starting at $min(D)$ and bin $b_k$ ending at $max(D)$.

One could argue, that choosing continuous distributions to model continuous features would be a superior choice and avoid the discretization and consequent loss of numeric precision.
Although it would have been possible to model the system using e.g. continuous normal distributions, we consciously decided against that, as a likely multimodal nature of the data would be lost.
More complex probability distributions that are able to model multimodality, e.g. Gaussian mixture models would be a solution to this,
but at the cost of requiring approximate inference methods for calculating the Bayesian Posterior.

In Bayesian probability theory, as described in Section \ref{subsec:bayesian_surprise}, the prior encodes one's beliefs in the distribution of interest, before any data is observed.
As such, the chosen prior probability distribution can have an important influence on the resulting model,
especially in the beginning, when the amount of observed data is still low.

As more data is collected and the prior gets repeatedly updated into the posterior (and is used as the new prior) during iterative Bayesian inference,
the effect of the specific choice of prior is reduced.
We chose to use a uniform prior over the $k$ bins per feature, i.e. we assume equal prior probability for all bins.

Every step, after Bayesian inference is performed for all features, and the posterior probabilities are calculated,
Bayesian Surprise is computed as the KL-divergence between the respective prior $P(M)$ and posterior $P(M|D)$: $D_{KL}(P(M|D)||P(M))$.
This way, surprise is calculated per feature per step on the trajectory.
In order to create a combined, single measure of surprise, a (weighted) sum of the separate results can be created. 

Besides the statistical modeling choices, another aspect that influences the resulting system is the stepsize,
i.e. the amount of measures performed per length of trajectory.
As every data point leads to an update of the model and the subsequent calculation of Bayesian Surprise,
this parameter has an effect on the behavior of the resulting model over time.
When a large stepsize is used, only few measurements along the trajectory will be performed.
Consequently, structural changes smaller than the stepsize (e.g. very small rooms or intersections of hallways) might be missed.
In the same way, a very small stepsize will over-condition the model on large amounts of only slightly varying measurements.

%% file: sections/evaluation.tex
\section{Evaluation}
\label{sec:evaluation}

This section first introduces the maps used in the creation and evaluation of the approach presented in this paper and also states the chosen parameters and design decisions. Afterwards, the system is intensely evaluated regarding several factors. We show that (1) the system reacts to unexpected events and also gets used to similar structures, (2) strong surprise does not discard the learned model, (3) the idea is transferable to real building plans, and (4) a traversed trajectory can be summarized using screenshots of the agent's local environment at places with high surprise.

\subsection{Evaluation Setup}
\label{subsec:evaluation_setup}
For the creation and evaluation of the approach presented in this paper, a total of 7 different maps was used.
Five of these are specially shaped, synthetic plans to show special behaviors of the algorithm, while two of the maps are based on plans of existing buildings.
Map \textbf{BasicSimple} represents the simplest scenario, namely a sequence of identically shaped rooms that just repeat.
The focus here is on the habituation on the map's structure and a surprise should only occur in special cases.
Maps \textbf{Alternating} and \textbf{AlternatingDoors} are quite similar to the previous map, but the rooms are not separated by doors,
but by differently shaped rooms. Thus, there are two different types of rooms that alternate.
Here it should be shown that on the one hand the two different types of rooms are recognized (visible by an increased surprise),
but on the other hand it is noticed that they repeat themselves.
The algorithm gets used to these alternating events.
Maps \textbf{AlternatingSurprise} and \textbf{AlternatingSurpriseDoors} are intended to provoke a habituation to given structures,
similar to the previous maps, but at one particular point there should be a sudden, strong surprise in form of a large room.
This should show that the strong surprise does not reject the learned concept of the environment.
Finally, two realistic maps are used in the course of this paper.
Map \textbf{LMU} presents a section of a university building of the University of Munich.
Map \textbf{TUM}, in contrast, models the main building of the Technical University of Munich on a larger scale,
i.e. a more complex map including an inner courtyard and also the four streets that run around the building.

The isovist generation framework was configured to use $360$ rays cast from the current vantage point of the agent for isovist calculation.
We found this value to be a good tradeoff between isovist accuracy and performance in most situations.
Nevertheless, in a limited amount of cases, measurement inaccuracies caused quite undesired effects in the model.
These cases will be discussed in more detail in the following sections.
For all evaluations performed, all features were modelled using Dirichlet Distributions with $K=10$.
Consequently, for each feature, the data was assigned to $10$ linearly spaced bins as described in Section \ref{subsec:calculation_of_bayesian_surprise}.
Furthermore, uniform priors were used that assign equal prior probability for all bins.
The stepsize was chosen such that the distance between two measurements on a straight trajectory is approximately 1 meter.
This parameter configuration was chosen as a tradeoff between the amount of model updates performed and the required size of structural changes impacting the model.
As stated in Section \ref{subsec:calculation_of_bayesian_surprise}, a large stepsize results in only few measurements along the trajectoy and will consequently 
miss out on structural changes. In the same way, a very short stepsize tends to over-condition the model very quickly on only slightly varying measurements.

\subsection{Habituation to Similar Structures}
\label{subsec:habituation}

An essential feature of Bayesian Surprise is that it reacts to unexpected events. Conversely, this means that more frequently observed events are correspondingly less surprising. In this section, we show that the system recognizes novel events, but quickly gets used to similar structures -- and also structural changes. A continuous descent of the surprise during a constant structure of the features is to be expected. Map \textbf{AlternatingDoors} is used for this part of the evaluation.

\begin{figure*}
	\centering
	\subfloat[area]{
		\includegraphics[width=.33\textwidth]{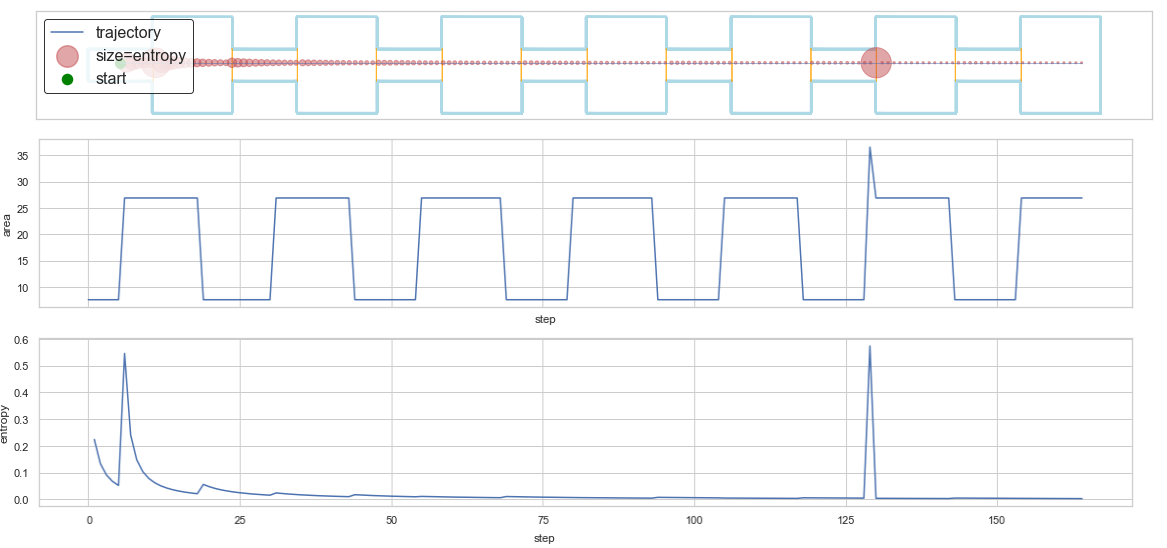}
		\label{fig:alternatingdoors-area-132}
	}
	\subfloat[real surface perimeter]{
		\includegraphics[width=.33\textwidth]{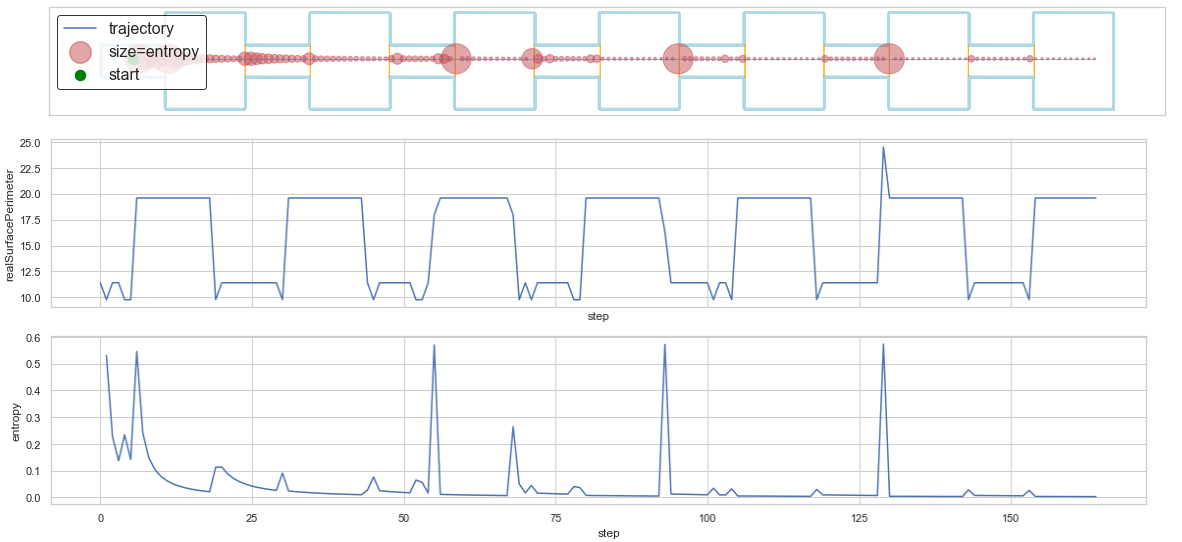}
		\label{fig:alternatingdoors-realsurfaceperimeter-130}
	}
	\subfloat[circularity]{
		\includegraphics[width=.33\textwidth]{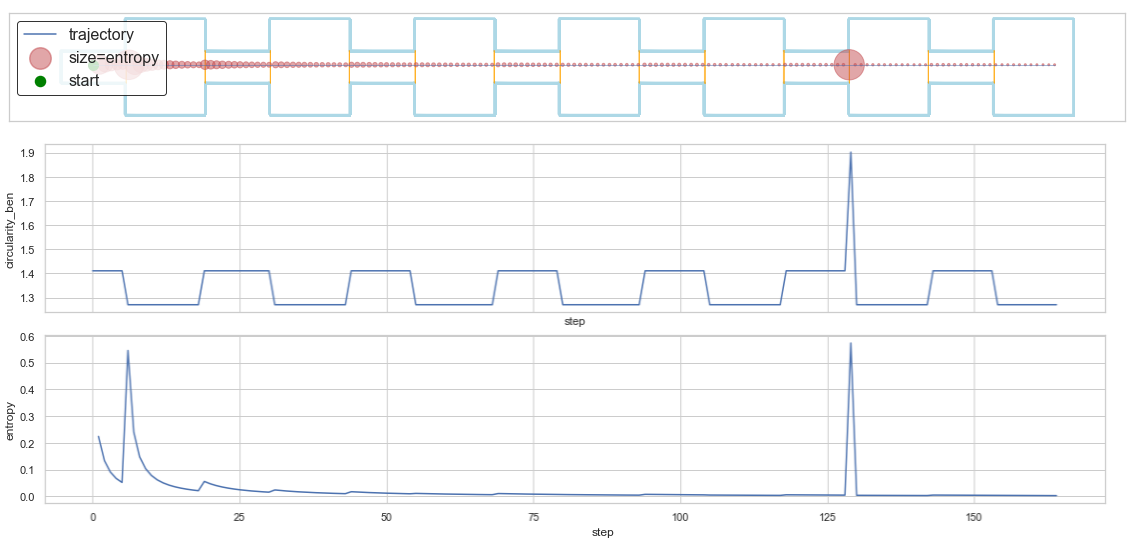}
		\label{fig:alternatingdoors-circularity-135}
	}
	\caption{Visualization of the calculated surprise based on individual features (\emph{area}, \emph{real surface perimeter}, \emph{circularity}) in map \textbf{AlternatingDoors}. Top: floor plan with marked surprise along traversed trajectory (from left to right). Middle: observed isovist measure. Bottom: calculated surprise in linear representation.} 
	\label{fig:habituation1}
\end{figure*}

The easiest to interpret feature is certainly \emph{area}. Since the map incorporates doors, there is only one fixed value of \emph{area} in each room: either there is lot of space to be seen or not (see Fig. \ref{fig:alternatingdoors-area-132} middle). The agent starts in the small room on the left-hand side in Fig. \ref{fig:alternatingdoors-area-132} top, and is moderately surprised by the observed impression (see Fig. \ref{fig:alternatingdoors-area-132} bottom for surprise in linear scale). This phenomenon will occur in all series of measurements since the models are initialized with a uniform distribution. The agent now moves through the small room, while the observed feature \emph{area} remains constant. This results in a decreasing surprise as the observed readings get more and more expected. When entering the first large room, the agent observes high values for \emph{area} that do not fit into the current model: the surprise is strongly increased. And again, the agent gets used to the current spatial impression while traversing the room, i.e. to the high values for \emph{area}. The entry into the second small room brings about a renewed, short-term increase of surprise. The reason for this is that the model has yet to ``level off''. From the second large room, however, the agent has ``understood'' the principle of the map and gets used to the structures. At the beginning of the next to last large room, there is an outlier in the observed readings (see Fig. \ref{fig:alternatingdoors-area-132} middle) and, accordingly, a strong peak in the measured surprise appears (Fig. \ref{fig:alternatingdoors-area-132} bottom). It can be seen that the measured feature value represents a global maximum, thus, this exact value has not yet occured. This is due to the way we approximate the isovist meaure \emph{area} by connecting the rays' endpoints and calculating the area of the resulting polygon, and also due to rounding errors. This amplitude does not occur regularly, as the used step length of the agent leads to the measurement of observations at different locations in different rooms.


The value curves of the observed features \emph{real surface perimeter} (see Fig. \ref{fig:alternatingdoors-realsurfaceperimeter-130} middle) and \emph{circularity} (Fig. \ref{fig:alternatingdoors-circularity-135} middle) are quite similar to that of \emph{area}. There is also a clear, alternating structure with high and low readings to be seen, but due to the approximation, the values of \emph{real surface perimeter} are not that binary. We also recognize an outlier in the course of the features. Looking at the surprise (Fig. \ref{fig:alternatingdoors-realsurfaceperimeter-130} bottom and Fig. \ref{fig:alternatingdoors-circularity-135} bottom), one can clearly see a continuous descent of the surprise, that is, a habituation to similar structures. The course of surprise with feature \emph{real surface perimeter} is not as stringent as in \emph{circularity}, since the observed feature values are likewise not.

\begin{figure*}
	\centering
	\subfloat[variance]{
		\includegraphics[width=.33\textwidth]{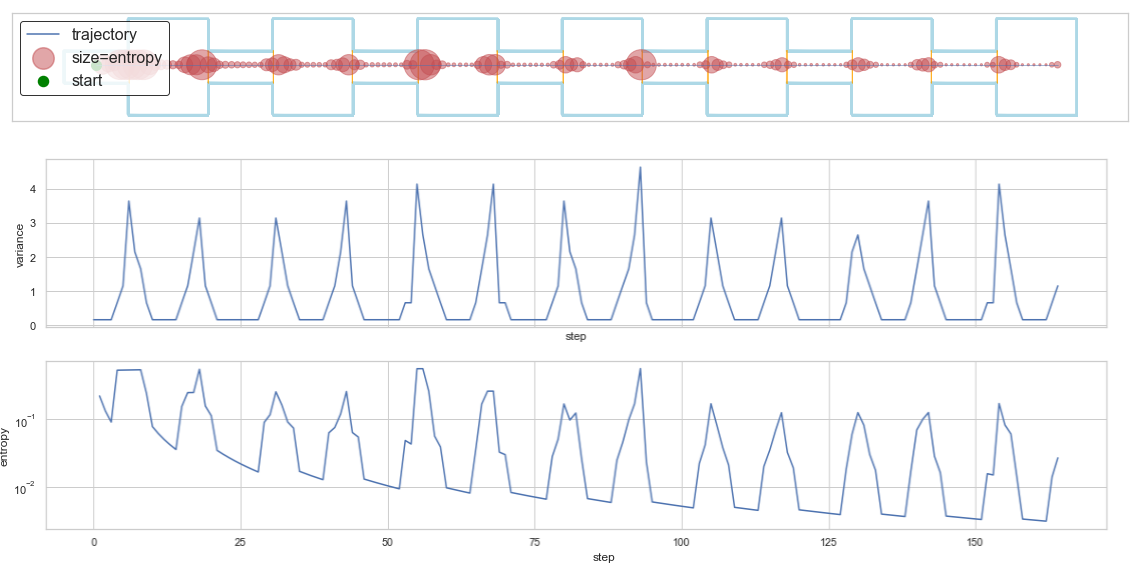}
		\label{fig:alternatingdoors-variance-133}
	}
	\subfloat[skewness]{
		\includegraphics[width=.33\textwidth]{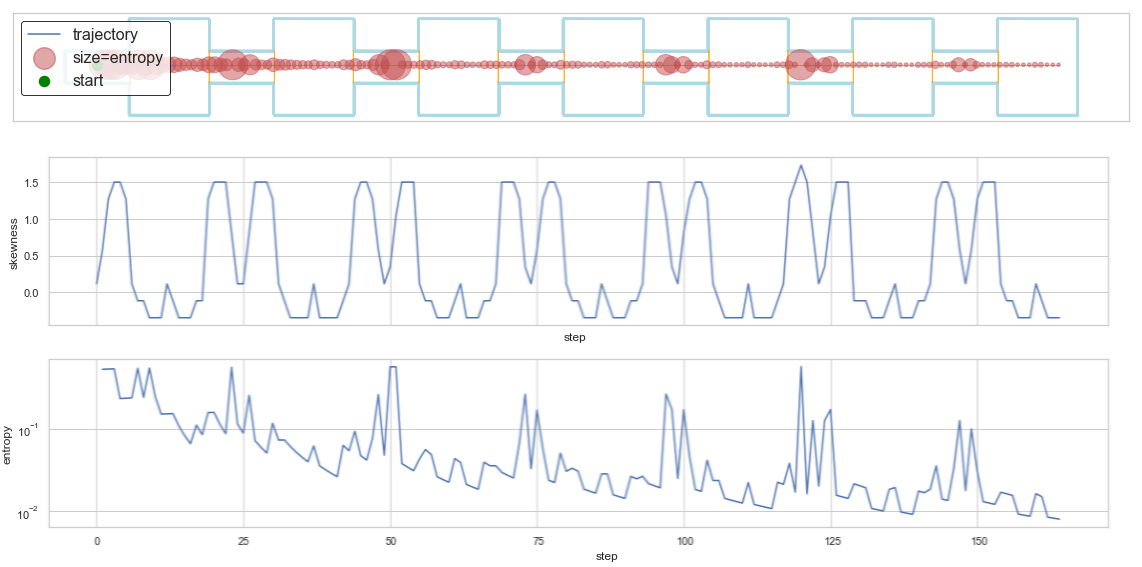}
		\label{fig:alternatingdoors-skewness-134}
	}
	\subfloat[occlusion]{
		\includegraphics[width=.33\textwidth]{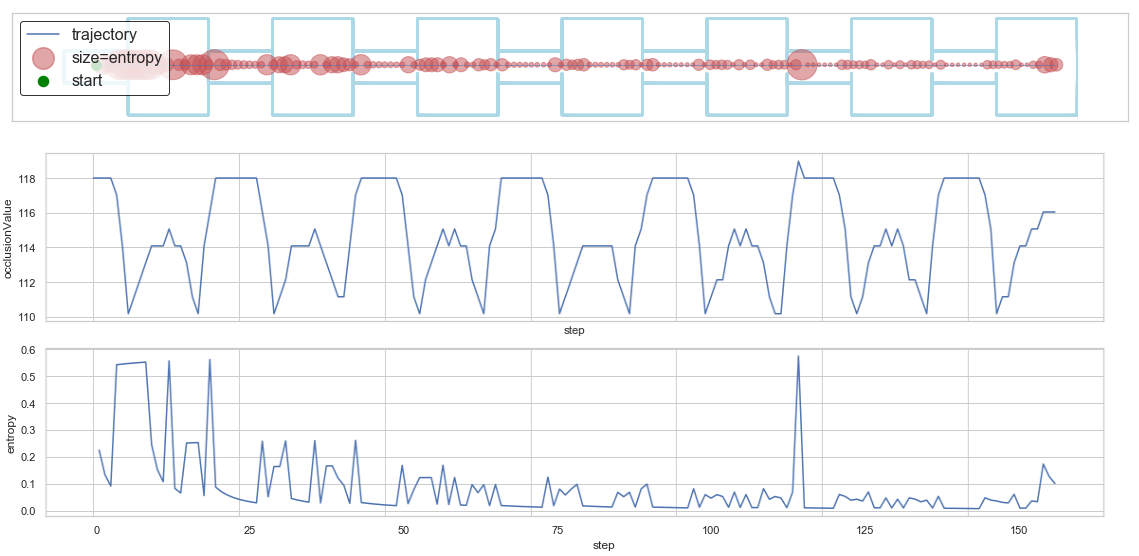}
		\label{fig:alternating-occlusion-125}
	}
	\caption{Visualization of the calculated surprise based on features \emph{variance}, \emph{skewness} and \emph{occlusion}. Please note that the surprise for \emph{variance} and \emph{skewness} is presented in logarithmic scale for map \textbf{AlternatingDoors}, while the surprise for \emph{occlusion} is presented in linear scale in map \textbf{Alternating}). Top: floor plan with marked surprise along traversed trajectory (from left to right). Middle: observed isovist measure. Bottom: calculated surprise.} 
	\label{fig:habituation2}
\end{figure*}

The measurements of \emph{variance} (see Fig. \ref{fig:alternatingdoors-variance-133} middle) and \emph{skewness} (Fig. \ref{fig:alternatingdoors-skewness-134} middle) are very similar. At the room boundaries we observe high values and low values in the middle of the rooms. In the case of \emph{skewness}, an additional local maximum is observed in the middle of the rooms. Due to the (intended) lack of synchronization of step lengths with the room sizes, it results that the observed values in different rooms are not identical. This is reflected in a somewhat increased surprise (see bottom of Fig. \ref{fig:alternatingdoors-variance-133} and Fig \ref{fig:alternatingdoors-skewness-134}, each in a logarithmic, not linear representation). Nonetheless, in both series of surprise a declining trend can be seen, indicating the habituation to similar structures.


Value \emph{occlusion} does not make sense in maps with doors, because there can be no masking or occlusion. For this reason, Fig. \ref{fig:alternating-occlusion-125} bottom shows the surprise for the observed \emph{occlusion} in the map \textbf{Alternating}, i.e. without doors. The value curve of \emph{occlusion} (Fig. \ref{fig:alternating-occlusion-125} middle) shows an alternating course, i.e. much obscured view when looking from a small room into a large room and correspondingly less obscuration if one looks from a large room into a small one. Due to missing doors, the readings are not that binary, which results in quite unsettled surprise values. In the linear course of the surprise (Fig. \ref{fig:alternating-occlusion-125} bottom), it can be seen, that the start in the small room and the traversion through the first large room provides many unfamiliar readings. From the second large room on, however, a concept has gradually been recognized and the surprise reduces continuously and remains so. Again, the observations contain a measurement error (next to last small room, see also global maximum in Fig. \ref{fig:alternating-occlusion-125} middle), which results in a short, strong surprise, but does not disturb the further course of surprise. In the last room the readings increase again, higher than in the small rooms but lower than in the large rooms. This is due to the fact that the last large room is indeed a large room, but has only one predecessor and no successor. Again, this is detected by the algorithm and signaled in surprise.

\begin{figure}
	\centering
	\includegraphics[width=\linewidth]{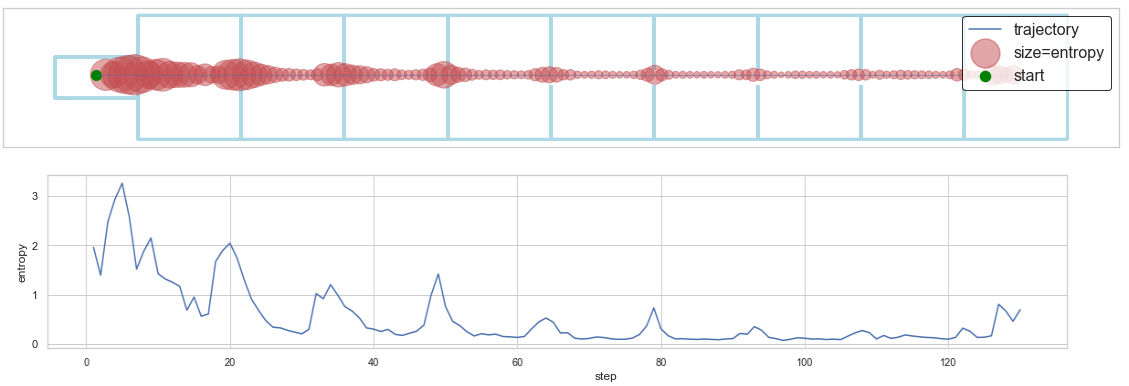}
	\caption{Additively combined surprise in map \textbf{BasicSimple}. The habituation of surprise to similar structure is visible.} 
	\label{fig:habituation3}
\end{figure}

Finally, Fig. \ref{fig:habituation3} shows the additively combined surprises of the six isovist measures in map \textbf{BasicSimple}, that is, without doors. It can be seen that the algorithm gets accustomed to the different value curves and briefly rises at the end, which is at least due to feature \emph{occlusion}, as already discussed.
Looking at the histograms of the calculated surprise values per isovist measure (not shown), it can be seen, that with one exception, all features contributed quite evenly to the surprise. Only the surprise histogram based on feature \emph{occlusion} reveals that more low surprise and less high surprise occured.

In summary, we can say that the algorithm is able to become accustomed to similar structures.

\subsection{Non-Rejection of Concept by Short Surprise}
\label{subsec:non_rejection_of_concept}

In the previous section we showed that the system recognizes structural changes, but is also able to get used to it. Now we show that a sudden, strong surprise does not discard the learned concept. For this purpose, we will discuss the feature \emph{area} in map \textbf{AlternatingSurpriseDoors} as an example. The use of doors keeps the feature values more consistent and is meant to demarcate the new environment more clearly. We also discuss feature \emph{occlusion} in map \textbf{AlternatingSurprise}, thus without doors, to show that a sudden surprise and the non-rejection of the concept is present even with a continuous change of the visual field.

\begin{figure*}
	\centering
	\subfloat[area]{
		\includegraphics[width=.5\textwidth]{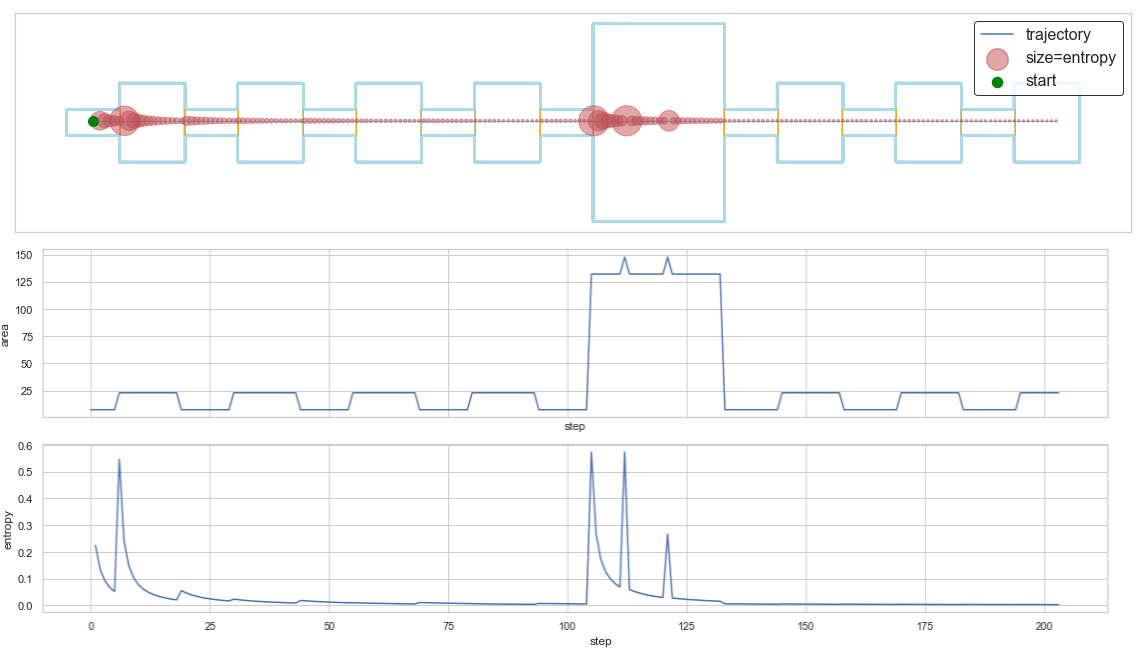}
		\label{fig:alternatingsurprisedoors-area-144}
	}
	\subfloat[occlusion]{
		\includegraphics[width=.5\textwidth]{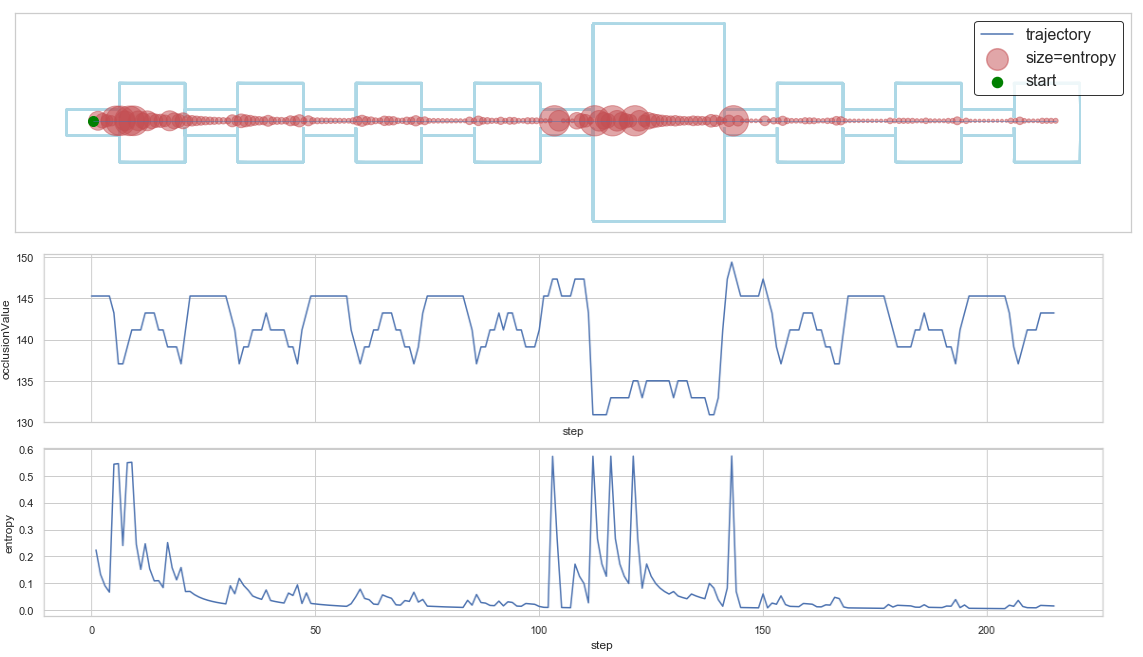}
		\label{fig:alternatingsurprise-occlusion-137}
	}
	\caption{Calculated surprise for features \emph{area} and \emph{occlusion} on maps \textbf{AlternatingSurpriseDoors} (\emph{area}) and \textbf{AlternatingSurprise} (\emph{occlusion}). Top: trajectory from left to right with highlighted surprise. Middle: observed isovist measure. Bottom: calculated surprise in linear scale.} 
	\label{fig:non-rejection}
\end{figure*}

As just described, due to doors, we obtain clearly separated readings for \emph{area} (see Fig. \ref{fig:alternatingsurprisedoors-area-144} middle), i.e. alternating low and medium values. Towards the middle of the map, the surprising, large room appears, which has correspondingly high readings. Once again we see measurement inaccuracies in the form of two peaks in the middle of the large room. After the novel room, the map follows the proven pattern again: small and medium sized rooms in turn. The structure of the map can be clearly seen in the course of surprise, see Fig. \ref{fig:alternatingsurprisedoors-area-144} bottom. At the very beginning of the agent's lifetime, as well as when entering the first middle-sized room, a strong surprise can be seen. However, this value decreases as expected and remains low. When entering the surprisingly large room, there occurs a strong discrepancy between expectation and observation, that is, the value of surprise is highly elevated. During the traversion of the large room, however, the surprise -- interrupted by measurement errors -- decreases. The most important finding here is the fact that when leaving the large room and entering the small room, the surprise performs a small jump down (see the visible bend in Fig. \ref{fig:alternatingsurprisedoors-area-144} bottom): thus, the algorithm recognizes the structure as it was before the surprising room.

Now we discuss feature \emph{occlusion} in a similar environment but without doors (doors would lead to no occlusion). The measured value (see Fig. \ref{fig:alternatingsurprise-occlusion-137} middle) is stable in the small rooms (much of the surrounding rooms are concealed) and in the medium-sized rooms it starts and ends with small values for \emph{occlusion}, while moderately much space of the smaller rooms is disguised when standing in the middle of the medium-sized rooms. Already in the last small room before the surprisingly large room, the structure of \emph{occlusion} changes. Now more area is masked, which results in increased readings and accordingly also in an increased surprise. After entering the large room, the readings go down and stay at a lower level, which can be read off again in surprise. The entry into the first small room after the uniquely large room is again unfamiliar, because at this point a value not yet measured occurs (see local maximum in Fig. \ref{fig:alternatingsurprise-occlusion-137} middle). Finally, it can be seen that from the end of the first small room after the surprising room, the system is again in a non-alarmed state.

In summary, it can be stated that the algorithm does not discard the learned concept when facing short, sudden surprises.

\subsection{Transferability to Real Building Plans}
\label{subsec:transferability_to_real}

After presenting the habituation to similar structures as well as the non-rejection of concept by sudden surprise with the help of synthetic maps, we discuss in this section the transferability to real building plans.

\begin{figure}[hpbt]
  \centering
  \includegraphics[width=\linewidth]{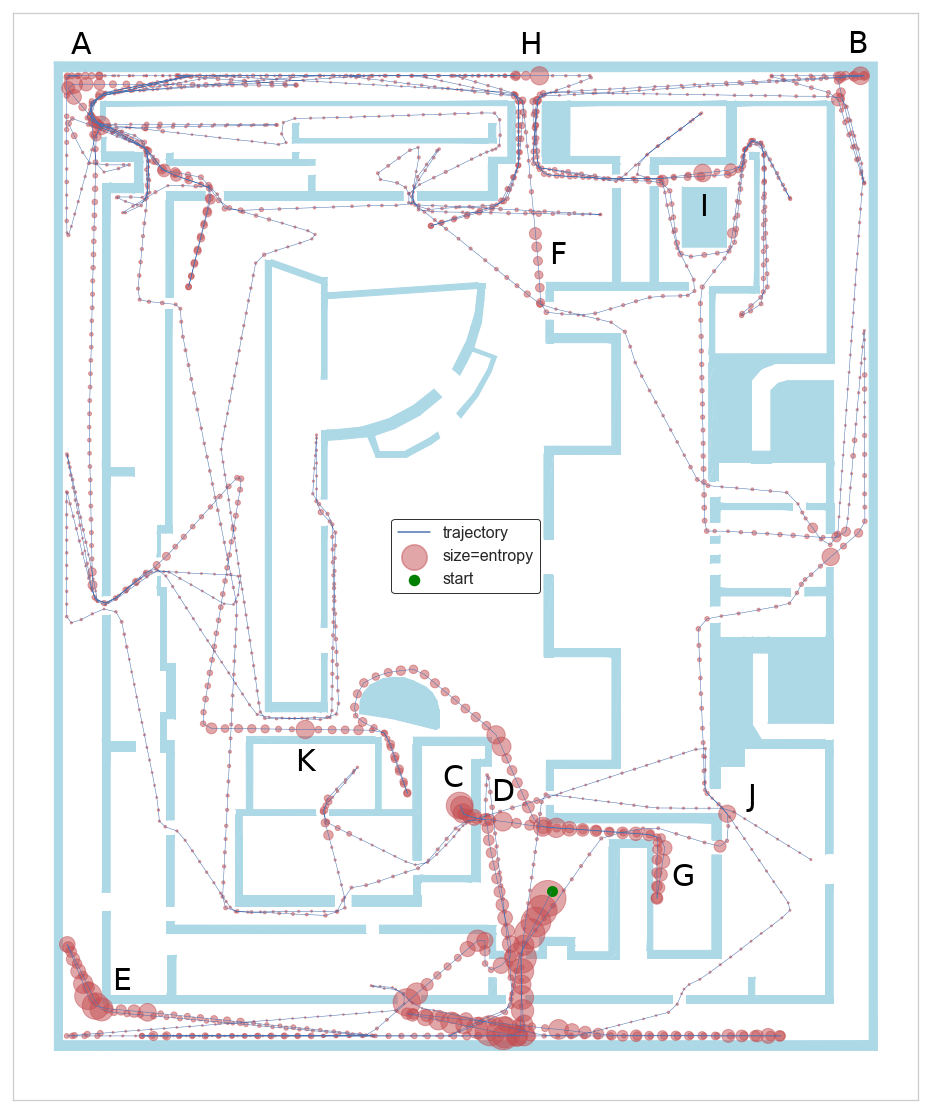}\\
  \caption{Floor plan \textbf{TUM} showing a very long trajectory through the building. The trajectory is highlighted with the calculated surprise based on all six isovist measures. The letters mark special regions where the surprise showed a remarkably high peak in at least one individual surprise. Those regions are interpreted in the text.} 
  \label{fig:tum-combined-158-annotated}
\end{figure}

Fig. \ref{fig:tum-combined-158-annotated} shows the building plan \textbf{TUM}. One can see the complex main building of the Technical University of Munich together with four adjoining streets. In the middle of the map is a large courtyard and the building has many differently shaped rooms with several entrances. The map shows trajectories together with the additive combined surprise based on all six isovist measures. Due to space restrictions we highlight only the most important and most piercing regions, which will be discussed below.

Based on feature \emph{real surface perimeter}, the strongest peaks of surprise are found at points \textbf{A}, \textbf{B} and \textbf{C}. \textbf{A} and \textbf{B} are exactly the locations where the agent can fully view both, the horizontal and vertical streets. These enormously high readings are remarkable and result in high values of surprise. Point \textbf{C}, however, is also responsible for a high surprise, but for a different reason: at this point, with regard to the agent's current model, the severely limited view was very surprising.

Points marked with \textbf{D} and \textbf{E} are based on particularly high surprise caused by measure \emph{occlusion}. Spot \textbf{D} is characterized by the fact that the agent steps out of smaller rooms into the front of the courtyard. A few steps after leaving the rooms, the view opens up and the projection of the part of the building just left obscures an enormous amount of space (top left of mark \textbf{D}). Similarly, albeit a little lower, the gaze goes down with concealment. Spot \textbf{E} identifies the agent's progress on the road below the building from right to left. The moment the agent walks around the corner, its field of view is obscured by the corner of the building wall: there is a high occlusivity. This surprise also withstands a few steps, but drops again after the agent goes back to the horizontal street.

Points \textbf{F} and \textbf{G} show sections of the route on which the value \emph{area} ensures a high surprise. In place \textbf{F}, one can clearly see the movement of the agent. The agent enters the courtyard from the building below spot \textbf{F} and receives a wide field of vision. This view does not remain steady high but grows even larger as the agent moves up and is able to look further into the left part of the map. Accordingly, a highly increased surprise can be measured here for a certain period of time. In place \textbf{G} there is also a movement to be seen. The agent moves from bottom to top, with the entrance on the right expanding the view and later, after entering the passage to the left, even more.

Point \textbf{H} marks a region where the readings for \emph{variance} lead to an increased surprise. At this point, the agent can look far to the left and right, but it can especially see the end of the courtyard below. Such a strong variance of the field of view has not yet come to the agent, thus, the corresponding surprise is very high.

Surprise based on the readings of \emph{skewness} is marked with \textbf{I} and \textbf{J}. In the small passage above spot \textbf{I} there seems to be a skew of the field of view, which is apparently novel given the current model of the agent. One can interpret this point in such a way that the field of vision is in most cases severely limited, only in one place the gaze seems to reach far, namely to the left through the passage and the door. Region \textbf{J} seems to be just as surprising. The interpretation is similar: close to the wall and just before the passage, the field of view is limited in many directions, but is in some places far (e.g. in the direction of the door on the right and in the direction of spot \textbf{D}).

Finally, spot \textbf{K} is an exemplary place where the \emph{circularity} measure has provided a peak in surprise. Again: based on the hitherto current model of the agent, the location seems to be special within the small corridor, here with a highly restricted vision up and down and a wide field of vision to the left and right.

In summary, the surprise, calculated on the given isovist measurements, makes the interesting regions of the complex map very well located. The algorithm finds street corners, views through the entire courtyard, leaving building parts into far-sighted areas and also very narrow and compact parts of the building. Noteworthy are also the places where the surprise even increases with the agent's movement.

\subsection{Characterizing a Trajectory}
\label{subsec:fingerprints_along_trajectory}

\begin{figure}
	\centering
	\subfloat[Hand selected trajectory.]{
		\includegraphics[width=.5\linewidth]{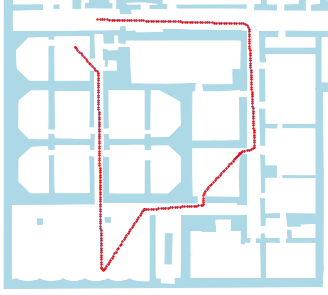}
		\label{fig:selection-1-traj}
	}
	\subfloat[Calculated surprise.]{
		\includegraphics[width=.5\linewidth]{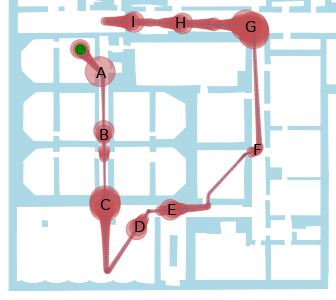}
		\label{fig:selection-1-map}
	}
	\\ 
	\subfloat[Measured area (top) and calculated surprise (bottom).]{
		\includegraphics[width=\linewidth]{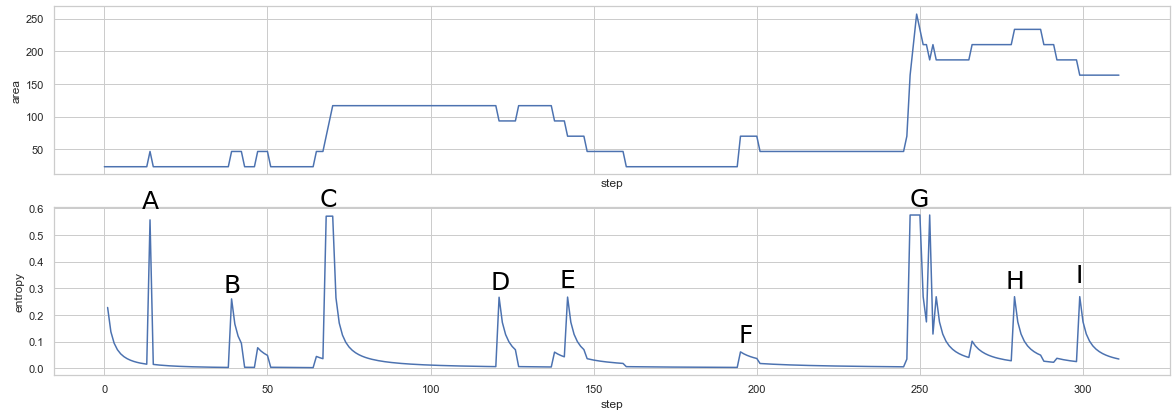}
		\label{fig:selection-1-curve}
	}
	\caption{Demonstration of the concept's feasibility in the sense of an application. (a) Detail of map \textbf{LMU} along with a hand selected trajectory that starts in the small room in the left-hand side of the map and runs counterclockwise. (b) Calculated surprise based on \emph{area} readings with peaks marked along the trajectory. (c) Observed \emph{area} readings along the trajectory (top) and corresponding calculated surprise in linear scale (bottom). These regions serve later as a visual summary of the trajectory traversed, see Fig. \ref{fig:selection-1-fingerprints}.}
	\label{fig:selection-1}
\end{figure}

After the transferability of the proposed concept to real floor plans was shown in the previous section, the concept's feasibility in the sense of an application will now be demonstrated. For this purpose, a hand-defined trajectory is selected in a cutout of map \textbf{LMU}. Along this trajectory the corresponding surprise is calculated. Fig. \ref{fig:selection-1-traj} shows the said trajectory starting in a room in the left-hand part of the map and runs counterclockwise.
For an easier interpretability, this section deals only with the observed readings for isovist measure \emph{area} and the corresponding surprise. Fig. \ref{fig:selection-1-curve} top shows the measured values for \emph{area}, while Fig. \ref{fig:selection-1-curve} bottom presents the resulting surprise in a linear scale together with highlighted peaks.
Correspondingly, Fig. \ref{fig:selection-1-map} depicts the cropped map with the trajectory together with the calculated surprises drawn as scaled circles as usual. The peaks marked in Fig. \ref{fig:selection-1-curve} correspond to the areas marked in Fig. \ref{fig:selection-1-map}.

Referring to the value development of feature \emph{area} (Fig. \ref{fig:selection-1-curve} top) and the resulting surprise (Fig. \ref{fig:selection-1-curve} bottom), it can clearly be seen that the agent starts in a smaller space and is surprised by the initial, novel ``sensory impression''. The surprise decreases as a result of constant valued observations. The surprise's first peak is at location \textbf{A}, where the agent leaves the room and enters the narrow, vertical hallway: the field of view is expanded, as the agent has not experienced that before. In place \textbf{B}, a surprise peak can be seen as the agent comes to a point of the corridor where two doors are exactly parallel. This constellation is also novel and thus surprises the agent. A very strong rise in the surprise is at point \textbf{C} where the agent enters the hall in the bottom of the map. Since in the following the measured \emph{area} remains at a constantly high level, the surprise accordingly decreases again. Point \textbf{D}, however, is still inside the hall, but here the point of view is obscured by the smaller obstacle a little further to the left. This fact can be seen in the observed readings as a local minimum. The agent now enters the narrow horizontal corridor (location \textbf{E}) and passes a door and a smaller room. Due to the applied binning of the isovist readings, the observed values seem to be constant (see Fig. \ref{fig:selection-1-curve} top) and there are no or just few small changes to the agent's model. A novel event can now be seen in place \textbf{F}, where the agent enters the longer vertical corridor: a slightly increased surprise is noted. The global maximum, both in the readings and in the surprise, can be found in spot \textbf{G}. The agent enters the very long horizontal corridor (only a section is visible, the map goes much further to the right) and the agent's field of view is very wide. In place \textbf{H} it is possible for the agent to look into the rooms above (cropped) and the surprise's last peak, at place \textbf{I}, can be found at a narrowing of the passage: this event is also novel.

\begin{figure}[hpbt]
  \centering
  \includegraphics[width=.8\linewidth]{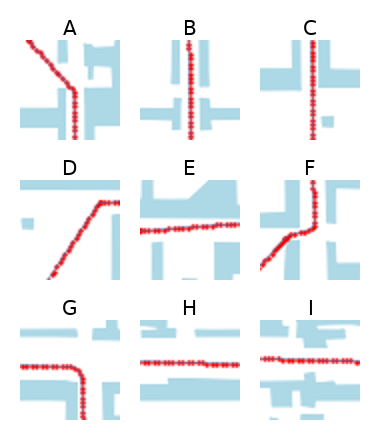}\\
  \caption{Screenshots of the agent's field of view in places with peaks in surprise. This list summarizes and characterizes the route traversed by the agent.}
  \label{fig:selection-1-fingerprints}
\end{figure}

If the task of an application is now to summarize or characterize the route run by the agent, an idea is to create some sort of screenshots of the agent's field of vision in the places where a high surprise prevailed. These screenshots were exemplary created using a squared clipping of the map exactly at points where the agent experienced peaks in surprise. Fig. \ref{fig:selection-1-fingerprints} shows nine screenshots labeled with the letters \textbf{A}-\textbf{I}. The characteristics of the route are clearly visible: the agent enters the vertical corridor (\textbf{A}), passes the parallel doors (\textbf{B}) and enters the hall (\textbf{C}). After going to the top right (\textbf{D}), the agent passes a horizontal narrow passage (\textbf{E}) and turns -- after some time -- to the right into a narrow vertical corridor (\textbf{F}). Afterwards, the agent turns left into a horizontal corridor (\textbf{G}), passing a door to the right-hand side while running to the left (\textbf{H}), and traverses a narrowing passage within the corridor (\textbf{I}).

In conclusion, one can see that a list of screenshots taken at regions with high surprise clearly summarizes the traversed trajectory.

%% file: sections/conclusion.tex
\section{Conclusion}
\label{sec:discussion}

With this paper we propose to combine the concepts of Bayesian Surprise and Isovist Analysis in the context of indoor navigation. Bayesian Surprise proposes that the subjective expectation of a person can be defined as the conditional probability distribution of events and that this distribution is constantly updated and corrected by new measurements. In our context, the observations are based on isovist measurements that are recorded along trajectories that traverse 2D floor plans.
The contribution of the paper is twofold. First, we present a way to generate isovist measures along spatial trajectories in a floorplan-based simulation environment in a way that can serve as input for the calculation of Bayesian Surprise. Second, we show how to compute Bayesian Surprise measures along the trajectories, that is, posterior inference and KL-divergence calculation.
The evaluation showed the following key aspects: (1) the system reacts to unexpected events along the trajectories and also gets used to similar structural changes, (2) strong surprise does not discard the model the agent learned, (3) our concept is transferable to floor plans of real buildings, and (4) a traversed trajectory can be summarized and characterized using screenshots of the agent's local environment taken at places with high surprise.

For future work, we plan to evaluate the effect of using continuous probability distributions instead of discrete ones.
Gaussian mixture models seem like a natural fit to capture the multimodal nature of the isovist measurements encountered along the indoor trajectories.
On one hand, this will increase the computational complexity, as approximate inference methods are required to estimate the Bayesian Posterior.
On the other hand, not needing to bin the data would allow for more gradual transitions.
This in turn could result in richer models, matching our human interpretation of surprise even better.